\documentclass[twocolumn]{el-author}   

\usepackage{color}
\usepackage{algorithmic,algorithm}

\newcommand{\ua}{\uparrow}
\newcommand{\nc}{\newcommand}
\nc{\da}{\downarrow} \nc{\hc}{\hat{c}} \nc{\hS}{\hat{S}}
\nc{\bra}{\langle} \nc{\ket}{\rangle} \nc{\eq}{equation (\ref}
\nc{\h}{\hat} \nc{\hT}{\h{T}}\nc{\be}{\begin{eqnarray}}
\nc{\ee}{\end{eqnarray}}\nc{\rd}{\textrm{d}}\nc{\e}{eqnarray}\nc{\hR}{\hat{R}}\nc{\Tr}{\mathrm{Tr}}
\nc{\tS}{\tilde{S}}\nc{\tr}{\mathrm{tr}}\nc{\8}{\infty}\nc{\lgs}{\bra\ua,\phi|}\nc{\rgs}{|\ua,\phi\ket}
\nc{\hU}{\hat{U}}\nc{\lfs}{\bra\phi|}\nc{\rfs}{|\phi\ket}\nc{\hZ}{\hat{Z}}\nc{\hd}{\hat{d}}\nc{\mD}{\mathcal{D}}
\nc{\bd}{\bar{d}}\nc{\bc}{\bar{c}}\nc{\mc}{\mathcal}\nc{\ea}{eqnarray}\nc{\mG}{\mathcal{G}}\nc{\bce}{\begin{center}}
\nc{\ece}{\end{center}}

\begin{document}

\title{A Robust Local Binary Similarity Pattern for Foreground Object  Detection}

\author{Dongdong Zeng, Ming Zhu and Hang Yang}

\abstract{Accurate and fast extraction of the foreground object  is one of the most significant issues to be solved due to its important meaning for object tracking and recognition in video surveillance.
Although many foreground object detection methods have been proposed in the recent past, it is still regarded as  a tough problem due to illumination variations and dynamic backgrounds challenges.
In this paper, we propose a robust foreground object  detection method with two aspects of contributions. First, we propose a robust texture operator named Robust Local Binary Similarity Pattern (RLBSP), which shows strong robustness to illumination variations and dynamic backgrounds.
Second, a combination of color and texture features  are  used to characterize pixel representations, which compensate each other to make full use of their own advantages.
Comprehensive experiments evaluated on the CDnet 2012 dataset demonstrate  that the proposed method    performs favorably against state-of-the-art methods.}

\maketitle

\section{Introduction}
Foreground object detection for a stationary camera is one of the essential tasks in  many computer vision and video analysis applications such as object tracking, activity recognition, and human-computer interactions.
As the first step in  these high-level operations, the accurate extraction of foreground objects directly affects the subsequent operations.
Background subtraction is the most  popular technology used for foreground object detection, the performance of foreground extract highly depended on the reliability of background modeling. In the past decades, various background subtraction methods have been proposed by researchers, most of them are classified as pixel-based method due to their low complexity and high processing speed. However, pixel-based methods are very sensitive to illumination variations in the scene, the remedy to this issue is to introduce the spatial information.
Local Binary Pattern (LBP) feature \cite{LBP} is one of the  earliest and most popular texture operator  proposed for background subtraction and has shown favourable performance due to its computational simplicity and tolerance against illumination variations. However, LBP is sensitivity to subtle local texture changes  caused by image noise.
To deal with this problem, Tan \textit{et al.}  proposed the Local Ternary Pattern (LTP) operator \cite{LTP} by adding a small tolerance offset. However, LTP is not invariant to illumination variations with scale transform  as the tolerance is constant for every pixel position.
Recently, Scale Invariant Local Ternary Pattern (SILTP) operator was proposed in \cite{SILTP} and shown its  tolerance against illumination variations and local image noise within a range,  due to employs  a scalable tolerance offset associated with the center pixel intensity.
More recently, Local Bianry Similarity Pattern (LBSP)  was proposed in \cite{LBSP}, this operator is based on absolute difference and is calculated  both inside a region of an image and across regions between two images to take more spatiotemporal information into consideration.
Despite the numerous texture features have been proposed, the main challenges in background modeling come from dynamic environments result from sudden or gradual illumination variations and periodic movements of backgrounds such as swaying trees, rippling water,  camera jitter, and so on. Thus, this task remains challenging when the scene contains illumination variations and dynamic backgrounds at the same time.
In this paper, we propose a novel texture feature named  Robust Local Binary Similarity Pattern (RLBSP),  which performs robustly against both illumination variations and dynamic backgrounds challenges.
Using the CDnet 2012 dataset \cite{CDnet2012}, we evaluate our method against numerous surveillance scenarios and the  experimental results show that the proposed method outperforms most state-of-the-art background subtraction methods.

\vspace{-0.02\linewidth}
\section{Robust Local Binary Similarity Pattern}\label{RobustLocalBinarySimilarityPattern}
\begin{figure}[t]
\centering{\includegraphics[width=1.0\linewidth]{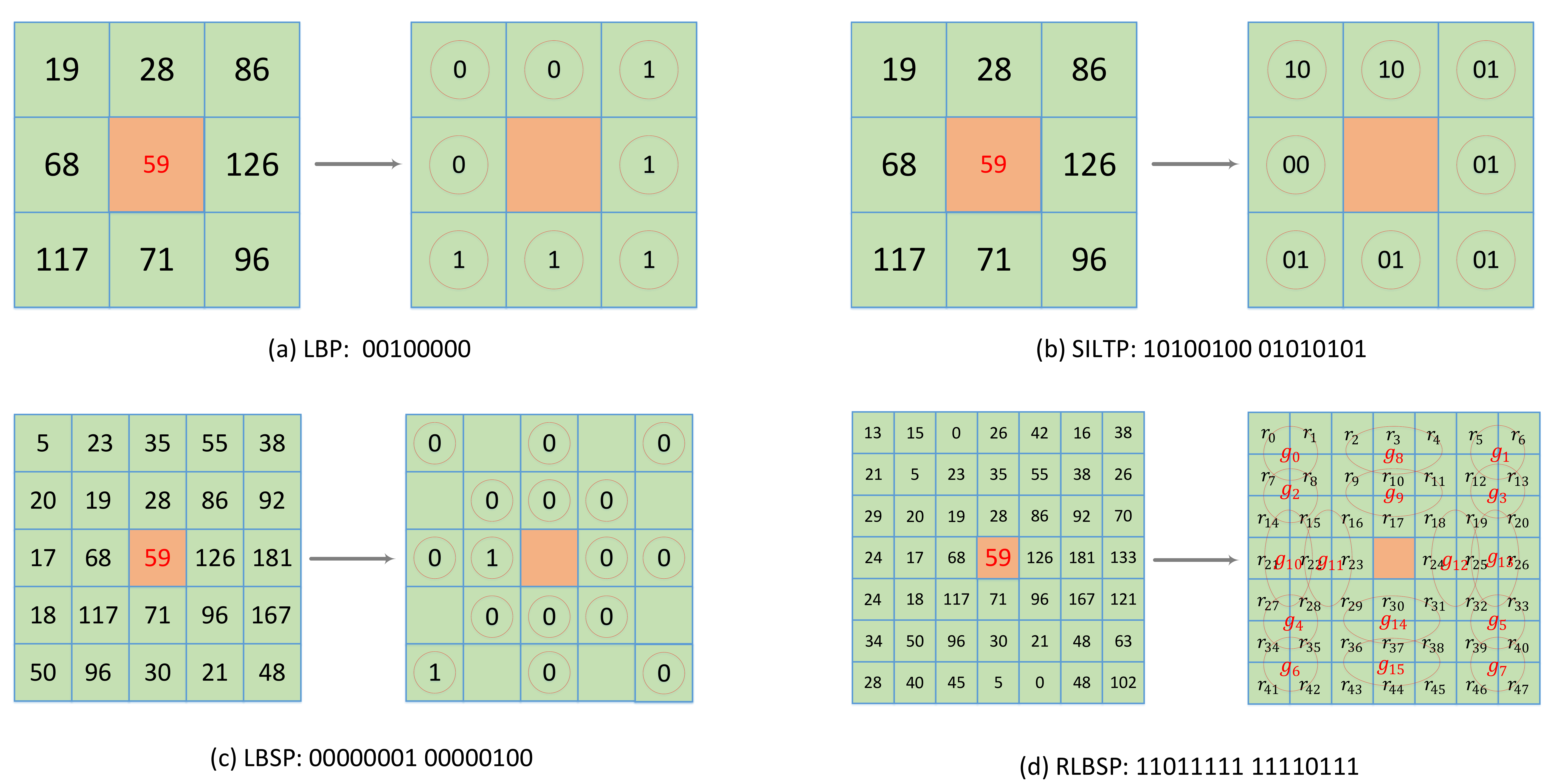}}
\vspace{-0.06\linewidth}
\caption{Examples of LBP \cite{LBP}, SILTP \cite{SILTP}, LBSP \cite{LBSP} and the proposed robust local binary similarity pattern (RLBSP).}
\label{RLBSP}
\end{figure}

Binary feature operators are popularly employed in background subtraction area due to their  low complexity,  discriminative  and  illumination invariance.
Most of them are based on the comparisons between pixel pairs in different configurations.
For example,  Local Binary Pattern (LBP)  works with eight-neighbors of each pixel, using the value of the center pixel as the threshold and considering the result as a binary string.
The Scale Invariant Local Ternary Pattern (SILTP) operator defines a new threshold strategy which makes the operator invariant under scale transform of pixel intensities, and each comparison will produce a two-bits encoder.
However, recent research  shows  that the two-bits encoding  is a coarse thresholding scheme due to the non-normal distribution of background statistics. Calculating the feature should not be based on strictly  comparisons such as $<$ and $>$,  but using absolute difference. Thus, in \cite{LBSP}, Local Bianry Similarity Pattern (LBSP) operator  proposes to calculate  local similarity in a $5\times5$ region based on the absolute difference.
Fig. \ref{RLBSP}(a$-$c) presents the encoding pattern of the three described texture operators.

However, these LBP-like features are often too sensitive to local changes in dynamic background and highly noise regions. To solve this problem, we propose a robust local binary similarity pattern operator called RLBSP.
Let a pixel $p$ at  a certain location of an image ${I}$, the coordinate of $p$ is $(x_p, y_p)$ and there are $P$  neighboring subregions $g_i$ spaced on an $n\times n$ region $R$ (see Fig. \ref{RLBSP}(d)). Then the RLBSP operator applied to $p$ can be expressed as follows:
\begin{equation}
\text{RLBSP}_{P,R}(p) =  \bigoplus_{i=0}^{P-1}S(I_{g_i}, I_p)
\end{equation}
where $I_p$ is the intensity value of center pixel $p$, $I_{g_i}$ is the average intensity value of its neighboring subregion $g_i$, $\bigoplus$ denotes concatenation operator of binary strings  and $S$ is a thresholding function which is defined as:
\begin{equation}
S(I_{g_i}, I_p)  =\begin{cases}
1, & \text{if } |I_{g_i} - I_p| \leq \tau I_p;\\
0, & \text{otherwise. }
\end{cases}
\label{RLBSPThreshold}
\end{equation}
where $\tau$ is a relative threshold value(which is set to 0.14 in this paper). The neighboring sub-areas $g_i$ are calculated as follows:
\begin{tiny}
\begin{equation}
\begin{cases}
g_0 = (r_{0}   + r_{1}  + r_{7}  + r_{8}     ) / 4;  g_8    = (r_{2} + r_{3} + r_{4} + r_{9} + r_{10} + r_{11}) / 6;\\
g_1 = (r_{5}   + r_{6}  + r_{12} + r_{13}    ) / 4;  g_9    = (r_{9} + r_{10} + r_{11} + r_{16} + r_{17} + r_{18}) / 6;\\
g_2 = (r_{7}   + r_{8}  + r_{14} + r_{15}    ) / 4;  g_{10} = (r_{14} + r_{15} + r_{21} + r_{22} + r_{27} + r_{28}) / 6;\\
g_3 = (r_{12}  + r_{13} + r_{19} + r_{20}    ) / 4;  g_{11} = (r_{15} + r_{16} + r_{22} + r_{23} + r_{28} + r_{29}) / 6;\\
g_4 = (r_{27}  + r_{28} + r_{34} + r_{35}    ) / 4;  g_{12} = (r_{18} + r_{19} + r_{24} + r_{25} + r_{31} + r_{32}) / 6;\\
g_5 = (r_{32}  + r_{33} + r_{39} + r_{40}    ) / 4;  g_{13} = (r_{19} + r_{20} + r_{25} + r_{26} + r_{32} + r_{33}) / 6;\\
g_6 = (r_{34}  + r_{35} + r_{41} + r_{42}    ) / 4;  g_{14} = (r_{29} + r_{30} + r_{31} + r_{36} + r_{37} + r_{38}) / 6;\\
g_7 = (r_{39}  + r_{40} + r_{46} + r_{47}    ) / 4;  g_{15} = (r_{36} + r_{37} + r_{38} + r_{43} + r_{44} + r_{45}) / 6.\\
\end{cases}
\label{subregion}
\end{equation}
\end{tiny}

Fig. \ref{RLBSP}(d) gives an example to show the encoding procedure of the proposed RLBSP texture operator. For a given pixel $p$ with an intensity value of 59 in the current frame $I$,
its $7\times 7$ neighboring region $R$ is divided into 16 subregions $g_i$. Then in each subregion $g_i$, we first calculate its average intensity value $I_{g_i}$ according to
Eq.(\ref{subregion}), and then make a comparison with the central pixel $I_p$ to get a one-bit encoder according to Eq.(\ref{RLBSPThreshold}). Finally, the 16 one-bit encoders are concatenated together to form a 16-bits string: 11011111 11110111, which is regarded as  the RLSBP operator of the given pixel $p$.

Since the RLBSP texture operator is an extension of other texture operators, it remains the advantages of the robustness to  illumination variations and local image noises. Besides, by taking  more spatial information into consideration and using the subregion average strategy, the proposed RLBSP is more robust for dynamic backgrounds and is more  promising for practical application.

\section{Modeling Background with RLBSP}\label{BackgroundModeling}
The local texture features achieve good tolerance against illumination variations in rich texture regions, however, it fails when both the foreground object and background image contain uniform and flat regions with less texture information. For example, walls or floors as background and color clothes as foreground have same texture information, so it is difficult to distinguish them by simply relying on the texture features.
To handle these situations, in this letter,  we proposed to create our background model  by integrating advantages of both texture feature and color feature.
Since texture features are not numerical values, traditional background modeling methods such as GMM \cite{gmm}, KED \cite{KDE} are not suitable for modeling RLBSP into background. Fortunately, inspired by the sample consensus method \cite{VIBE}, we develop an innovative mechanism that combines RLBSP feature and color feature.
The overview of the procedure is presented in \textbf{Algorithm} \ref{algorithmProcess}. The algorithm contains several parameters and their values are similar to \cite{VIBE} with: $N=50$, $\#_{min}=2$, $R_c = 15$, $R_t = 5$, and $\phi = 16$.

\vspace{-0.02\linewidth}
\begin{algorithm}[ht]
\caption{: Foreground Object Detection Procedure}
\begin{algorithmic}[1]
\STATE {Initialization}.\\
\textbf{for} each pixel $p(x,y)$ in the first frame $I^{0}$ \textbf{do}\\
\quad extract color feature $I(x,y)$ and texture feature ${RLBSP}(x,y)$ \\
\quad construct the background sample of $p(x,y)$ with: \\
\quad $F(x,y) = \{I(x,y), RLBSP(x,y)\}$ \\
\quad construct the background model of $p(x,y)$ with $N$ random neighboring background samples:\\
\quad \textbf{for} $i = 1,\ldots,N$ \textbf{do}\\
\qquad $\mathcal{B}_i(x,y) = \{ F(\bar{x}, \bar{y}  )    | (\bar{x}, \bar{y}) \in \mathcal{N}(x,y)             \}$\\
\quad \textbf{end for}\\
\textbf{end for}\\
---------------------------------------------------------------------------------
\STATE {Foreground Detection}.\\
\textbf{for} each pixel $p(x,y)$ in the current frame $I$ \textbf{do}\\
\quad extract color feature $I(x,y)$ and texture feature ${RLBSP}(x,y)$\\
\quad $nMatches$ = $0$, $i$ = $ 0$\\
\quad \textbf{while} $nMathes <  \#_{min}$  and $ i < N$\\
\qquad \quad $colorDist = L_{1}Dist( I(x,y), I_i(x,y)) $\\
\qquad \quad \textbf{if} $colorDist \geq R_c$\\
\qquad \quad \quad \textbf{goto} \textbf{failedMatch};\\
\qquad \quad $textDist = HamDist( RLBSP(x,y), RLBSP_i(x,y)) $\\
\qquad \quad \textbf{if} $textDist \geq R_{t}$\\
\qquad \quad \quad \textbf{goto} \textbf{failedMatch};\\
\qquad \quad  $nMatches$++;\\
\qquad \quad \textbf{failedMatch}:\\
\qquad \quad \quad $i$++;\\
\quad \textbf{end while}\\
\quad \textbf{if} $nMatches < \#_{min}$\\
\qquad \quad $p(x,y) $ is foreground;\\
\quad \textbf{else}\\
\qquad \quad $p(x,y)$ is background; \\
\textbf{end for}\\
---------------------------------------------------------------------------------
\STATE {Background Model Update}. \\
\textbf{for} each pixel $p(x,y)$ in the current frame $I$ \textbf{do}\\
\qquad \textbf{if} $p(x,y)$ is background   \\
\qquad \qquad \textbf{if} $rand()$  \%  $\phi == 0$   \\
\qquad \qquad \qquad update $\mathcal{B}({x},{y})$ with $F(x, y)$\\
\qquad \qquad \textbf{if} $rand()$  \%  $\phi == 0$   \\
\qquad \qquad \qquad update  $\mathcal{B}(\bar{x},\bar{y})$ with $F(x, y)$\\
\qquad \textbf{else} \\
\qquad \qquad \textbf{return}\\
\textbf{end for}
\end{algorithmic}
\label{algorithmProcess}
\end{algorithm}

\vspace{-0.08\linewidth}
\section{Experimental results}
To evaluate the use of RLBSP for foreground object detection, we use  a standard  Change Detection dataset (CDnet 2012) available in \cite{CDnet2012}. Seven different quantitative metrics have been defined: Recall (\textit{Re}), Specificity (\textit{Sp}), False positive rate (\textit{FPR}), False negative rate (\textit{FNR}), Percentage of wrong classifications (\textit{PWC}), Precision (\textit{Pr}), F-Measure (\textit{FM}). All these metrics  range  from 0 to 1,  for \textit{Re, Sp, Pr} and \textit{FM} metrics, higher values indicate more accuracy while for \textit{PWC, FNR} and \textit{FPR} metrics, lower values indicate better performance.  Generally speaking, a foreground detection method is considered good if it gets high recall scores and without sacrificing precision. So, the F-Measure metric is mainly  accepted as a good indicator of the overall performance of  different methods. We use the tools provided by the authors of \cite{CDnet2012} to compute these metrics. In Table \ref{resultsofCDnet2012}, we report the  seven metric scores  on each category in CDnet 2012 dataset. We can see that our method performs well on the \textit{dynamic background, camera jitter} and \textit{shadow} categories.
Then, we compare the F-Measure performance of the proposed method against some state-of-the-art methods in  Table \ref{percategoryFM}, the results show that our RLBSP-based foreground detection method is much more robust than other methods with LBSP \cite{LBSP} and \cite{LBP} texture features on the \textit{dynamic background} and \textit{camera jitter} categories. The processing speed of the proposed method is also evaluated to show that it has low computational complexity and is suitable for real time applications. For a  sequence with the frame size of $320 \times 240$, it runs at 41 fps with a 3.6 GHz Intel Core i7 7700  CPU.
\begin{table}[t]
\begin{scriptsize}
\begin{center}
\caption{Complete results obtained by RLBSP  on the CDnet 2012 dataset}
{\tabcolsep3pt\begin{tabular}{cccccccc} \hline
  Category&  Recall&  Specificity&  FPR&  FNR&  PWC&  Precision&  F-Measure\\
\hline
baseline          & 0.9409   & 0.9973  & 0.0027  & 0.0591  & 0.4842  & 0.9144  & 0.9272\\
cameraJ           & 0.8878   & 0.9775  & 0.0225  & 0.1122  & 2.6135  & 0.7290  & 0.7863\\
dynamic           & 0.9068   & 0.9941  & 0.0059  & 0.0932  & 0.6788  & 0.6754  & 0.7185\\
intermittent      & 0.7065   & 0.9294  & 0.0706  & 0.2935  & 8.4502  & 0.6263  & 0.6138\\
shadow            & 0.9343   & 0.9911  & 0.0089  & 0.0657  & 1.1257  & 0.8437  & 0.8850\\
thermal           & 0.7863   & 0.9945  & 0.0055  & 0.2137  & 1.3863  & 0.8748  & 0.8117\\
\hline
Overall           &0.8604    & 0.9807  & 0.0193  & 0.1396  & 2.4564  & 0.7773  & 0.7904\\
\hline
\end{tabular}}{}
\label{resultsofCDnet2012}
\end{center}
\end{scriptsize}
\end{table}
\hspace{-0.1\linewidth}
\begin{table}[t]
\begin{scriptsize}
\begin{center}
\caption{Per-category and overall  F-Measure scores by different methods }
{\tabcolsep3pt\begin{tabular}{ccccccc|c} \hline
Method   &  Baseline&  CameraJ&  Dynamic&  Intermittent&  Shadow&  Thermal    &   Overall    \\
        &  FM      &  FM             &  FM                &  FM             &  FM             &  FM           &  FM  \\
\hline
{{\color{red}{RLBSP}}}          &   {0.9272}&\bf{0.7863}&\bf{0.7185}&\bf{0.6138}&\bf{0.8850}&\bf{0.8117}&\bf{0.7904}\\
LBSP \cite{LBSP}                &\bf0.9320  & 0.7462    & 0.5664    & 0.5940    & 0.8696    & {0.7803}  &	0.7481\\
LBP \cite{MultiLayer}    & 0.9004    & 0.7311    &   {0.6278}& 0.4816    & 0.8099    & {0.6331}  &	0.6993\\
{KDE  \cite{KDE}}               &   {0.9092}& {0.5720}  &   {0.5961}&   {0.4088}&   {0.7660}&   {0.7423}&   {0.6719}\\
{GMM  \cite{gmm}}               &   {0.8245}&   {0.5969}&   {0.6330}&   {0.5207}&   {0.7156}&   {0.6621}&   {0.6624}\\
\hline
\end{tabular}}{}
\label{percategoryFM}
\end{center}
\end{scriptsize}
\end{table}

\section{Conclusion}\label{CONCLUSION}
In this paper, we propose a robust texture operator named Robust Local Binary Similarity Pattern (RLBSP) for  foreground object detection. The RLBSP shows strong robustness to illumination variations and dynamic backgrounds scene conditions.  To handle the limitation   of texture feature in  uniform regions, we  combine the color and texture features   to characterize pixel representations.
Experiments evaluated on the CDnet 2012 dataset show that the proposed method outperforms  state-of-the-art methods and runs in real-time performance.

\vskip3pt
\ack{This research is supported by the National Science Foundation of China under Grant No.61401425.}

\vskip5pt


\noindent E-mail: zengdongdong13@mails.ucas.edu.cn

%
%
%
%


\begin{thebibliography}{1}
\providecommand{\url}[1]{#1}
\csname url@samestyle\endcsname
\providecommand{\newblock}{\relax}
\providecommand{\bibinfo}[2]{#2}
\providecommand{\BIBentrySTDinterwordspacing}{\spaceskip=0pt\relax}
\providecommand{\BIBentryALTinterwordstretchfactor}{4}
\providecommand{\BIBentryALTinterwordspacing}{\spaceskip=\fontdimen2\font plus
\BIBentryALTinterwordstretchfactor\fontdimen3\font minus
  \fontdimen4\font\relax}
\providecommand{\BIBforeignlanguage}[2]{{%
\expandafter\ifx\csname l@#1\endcsname\relax
\typeout{** WARNING: IEEEtran.bst: No hyphenation pattern has been}%
\typeout{** loaded for the language `#1'. Using the pattern for}%
\typeout{** the default language instead.}%
\else
\language=\csname l@#1\endcsname
\fi
#2}}
\providecommand{\BIBdecl}{\relax}
\BIBdecl

\bibitem{LBP}
M.~Heikkila and M.~Pietikainen, ``A texture-based method for modeling the
  background and detecting moving objects,'' \emph{{IEEE} Trans. Pattern Anal.
  Mach. Intell.}, vol.~28, no.~4, pp. 657--662, 2006.

\bibitem{LTP}
X.~Tan and B.~Triggs, ``Enhanced local texture feature sets for face
  recognition under difficult lighting conditions,'' \emph{{IEEE} Trans. Image
  Process.}, vol.~19, no.~6, pp. 1635--1650, 2010.

\bibitem{SILTP}
S.~Liao, G.~Zhao, V.~Kellokumpu, M.~Pietik{\"a}inen, and S.~Z. Li, ``Modeling
  pixel process with scale invariant local patterns for background subtraction
  in complex scenes,'' in \emph{Proc. IEEE Conf. Comput. Vis. Pattern
  Recognit.}, 2010, pp. 1301--1306.

\bibitem{LBSP}
P.-L. St-Charles and G.-A. Bilodeau, ``Improving background subtraction using
  local binary similarity patterns,'' in \emph{Proc. IEEE Winter Conf. Appl.
  Comput. Vision.}, 2014, pp. 509--515.

\bibitem{CDnet2012}
N.~Goyette, P.-M. Jodoin, F.~Porikli, J.~Konrad, and P.~Ishwar,
  ``Changedetection. net: {A} new change detection benchmark dataset,'' in
  \emph{Proc. IEEE Conf. Comput. Vis. Pattern Recognit. Workshops}, 2012, pp.
  1--8.

\bibitem{gmm}
C.~Stauffer and W.~E.~L. Grimson, ``Adaptive background mixture models for
  real-time tracking,'' in \emph{Proc. IEEE Conf. Comput. Vis. Pattern
  Recognit.}, 1999, pp. 246--252.

\bibitem{KDE}
A.~Elgammal, R.~Duraiswami, D.~Harwood, and L.~S. Davis, ``Background and
  foreground modeling using nonparametric kernel density estimation for visual
  surveillance,'' \emph{Proc. {IEEE}}, vol.~90, no.~7, pp. 1151--1163, 2002.

\bibitem{VIBE}
O.~Barnich and M.~Van~Droogenbroeck, ``{ViBe}: A universal background
  subtraction algorithm for video sequences,'' \emph{{IEEE} Trans. Image
  Process.}, vol.~20, no.~6, pp. 1709--1724, 2011.

\bibitem{MultiLayer}
J.~Yao and J.-M. Odobez, ``Multi-layer background subtraction based on color
  and texture,'' in \emph{Proc. IEEE Conf. Comput. Vis. Pattern Recognit.},
  2007, pp. 1--8.

\end{thebibliography}
\end{document}